\documentclass[runningheads]{llncs}
\usepackage[T1]{fontenc}
\usepackage{graphicx}
\usepackage{geometry}
\usepackage{xspace}
\usepackage{amsfonts,amsmath,amssymb}
\usepackage{xcolor}
\usepackage{tikz}
\usepackage{graphicx}
\usepackage{tabularx}
\usepackage{array}
\usepackage{multirow}
\usepackage{caption}  
\usepackage{subcaption}
\usepackage{hyperref, natbib}
\usepackage{expex}
\lingset{everygla=\upshape}  

\begin{document}
\title{Quantum NLP models on Natural Language Inference}
%
%
\author{Ling Sun\orcidID{0009-0004-7256-2544} \and
Peter Sullivan \and
Michael Martin \and
Yun Zhou} 
\authorrunning{L. Sun et al.}
%
\institute{Indiana University, Bloomington IN 47405, USA}

\maketitle              
\begin{abstract}
Quantum natural language processing (QNLP) offers a novel approach to semantic modeling by embedding compositional structure directly into quantum circuits. This paper investigates the application of QNLP models to the task of Natural Language Inference (NLI), comparing quantum, hybrid, and classical transformer-based models under a constrained few-shot setting. Using the lambeq library and the DisCoCat framework, we construct parameterized quantum circuits for sentence pairs and train them for both semantic relatedness and inference classification. To assess efficiency, we introduce a novel information-theoretic metric—Information Gain per Parameter (IGPP)—which quantifies learning dynamics independent of model size. Our results demonstrate that quantum models achieve performance comparable to classical baselines while operating with dramatically fewer parameters. The Quantum-based models outperform randomly initialized transformers in inference and achieve lower test error on relatedness tasks. Moreover, quantum models exhibit significantly higher per-parameter learning efficiency—up to five orders of magnitude more than classical counterparts—highlighting the promise of QNLP in low-resource, structure-sensitive settings. To address circuit-level isolation and promote parameter sharing, we also propose a novel cluster-based architecture that improves generalization by tying gate parameters to learned word clusters rather than individual tokens.

\end{abstract}
\section{Introduction}
Natural Language Inference (NLI) is a fundamental challenge in natural language processing, requiring models to determine whether one sentence logically follows from, contradicts, or is unrelated to another \citep[e.g.,][]{hu2019monalog, richardson2020probing}. Solving NLI demands not just lexical understanding but also compositional reasoning—the ability to understand how meanings of phrases combine.

Inference is inherently compositional, yet most classical models—especially transformers—lack explicit mechanisms to model compositional structure. While it is possible to improve classical models by augmenting them with annotated data that captures world knowledge and compositional semantics, this process is resource-intensive and does not guarantee generalization. Quantum circuit models, in contrast, offer a novel approach: by leveraging compositional frameworks such as DisCoCat \citep{coecke2010mathematical, bradley2018translating, miranda2022quantum}, they natively encode syntactic and semantic structure into circuit form. Yet, the application of these models to practical language tasks remains underexplored.

In this work, we introduce a few-shot benchmark for quantum NLI, designed to reflect the constraints of near-term quantum devices.\footnote{Our codes is available at \url{https://github.com/LingSyrina/QNLP}.} We limit both data and circuit size—training on only 100 sentence pairs using 1 qubit per word in classical simulation—to assess whether quantum models can learn structured meaning efficiently under resource constraints (Section~\ref{experiment}).

We propose and evaluate a new information-theoretic metric, Information Gain per Parameter (IGPP), to quantify learning efficiency independent of model size. Our experiments show that quantum models, even in such constrained settings, demonstrate competitive inference performance and dramatically higher learning efficiency, outperforming random-initialized transformers and approaching pretrained baselines (Section~\ref{results}).

However, we identify a key obstacle to quantum generalization: circuit isolation, where each unique input generates a distinct, unshared circuit. To address this, we introduce a novel Cluster model that groups similar circuit boxes and promotes parameter sharing. Our preliminary results show this approach improves macro F1 and mitigates isolation effects (Section~\ref{cluster}).

Finally, we discuss the limits of current quantum models in generalizing to word- and composition-level patterns, pointing toward future work in shared parameterization and scalable architectures (Sections~\ref{discussion}-\ref{limitation}).


\section{Preliminaries and Previous Work}
NLI tasks require models to determine the logical relationship between sentences. Successful approaches often rely not only on world knowledge but also on an understanding of compositional structures (see Section~\ref{NLI}). Quantum circuit models, such as those based on the DisCoCat framework \citep{coecke2010mathematical}, offer promising capabilities for naturally integrating compositional and syntactic information relevant to NLI tasks (see Section~\ref{circuit}). However, this area remains largely underexplored.

\subsection{Natural language inference task} \label{NLI}

NLI is the task of determining the logical relationship between two sentences. For example, given the premise “Mary ate an apple,” one can infer that “Mary ate some fruit” is also true. This type of relationship is known as entailment. A common diagnostic in semantics is the cancellation test: native speakers typically judge the sentence “Mary ate an apple but did not eat any fruit” as contradictory, since the first clause entails the second. Most datasets designed for NLI tasks include three core types of sentence pair relations \citep{maccartney-manning-2008-modeling}: entailment (\ref{relation_a}), neutral (\ref{relation_b}), and contradiction (\ref{relation_c}). Importantly, these relations are not necessarily symmetric. As illustrated in (\ref{relation_a}), “The cat is sleeping on the mat” entails “The cat is on the mat,” but not vice versa. If “The cat is on the mat” serves as the premise, then “The cat is sleeping on the mat” may be true or false, resulting in a neutral relationship. 

\begin{enumerate}
    \item[a] \textbf{Entailment} \label{relation_a} \\
    “The cat is sleeping on the mat.” \textit{entails} “The cat is on the mat.”

    \item[b] \textbf{Neutral} \label{relation_b} \\
    “The man is eating an apple.” is \textit{neutral} with respect to “The man is sitting in a park.”

    \item[c] \textbf{Contradiction} \label{relation_c} \\
    “The dog is barking loudly.” \textit{contradicts} “The dog is completely silent.”
\end{enumerate}

NLI has long been regarded as a fundamental task in NLP, attracting significant research interest \citep[e.g.,][]{hu2019monalog, richardson2020probing}. This is not only due to its relevance for downstream applications such as question answering \citep{demszky2018transforming}, dialogue systems \citep{dziri2019evaluating}, and information retrieval \citep{samarinas2020latent}, but also because it poses unique challenges to traditional generative and encoder–decoder models. Unlike tasks that rely primarily on sentential syntax or distributional semantics, NLI demands both lexical knowledge and the ability to perform complex compositional reasoning. For example, to correctly infer the relationship between “Mary ate an apple” and “Mary ate some fruit,” a model must recognize that an apple is a type of fruit—a fact that requires external commonsense world knowledge. Furthermore, understanding how phrases compose semantically is crucial. The difference between “some apples” and “all apples” illustrates how quantifiers affect inference: “some apples” entails “some fruits,” while “all fruits” entails “all apples”. The quantifiers “some” and “all” influence the directionality of entailment, demonstrating the need for reasoning over compositional structure.

Many of the most effective models in NLI incorporate world knowledge and compositional reasoning by leveraging grammatical annotations such as categorical grammar \citep{bernardi2007lite}, dependency structures \citep{yusuf2017basic}, and monotonicity information \citep{hu2019monalog, chen2021neurallog}. One notable example is the work by \citet{chen2021neurallog}, which utilizes monotonicity annotations to implement compositional inference rules, supported by a word-level monotonicity corpus. Their best-performing model achieved an accuracy of 0.903 on the Sentences Involving Compositional Knowledge (SICK) dataset \citep{marelli-etal-2014-sick}, outperforming other models. This result highlights the importance of incorporating semantic and logical structures beyond surface-level syntax and lexical patterns in NLI systems.

\subsection{Categorical Compositional Distributional framework}\label{circuit} 
The Categorical Compositional Distributional framework (DisCoCat) \citep{coecke2010mathematical} offers a principled approach to embedding linguistic structure directly into model architecture. Grounded in category theory, DisCoCat distinguishes between words based on their syntactic roles—for example, treating nouns and verbs as different types of tensors or quantum states. This allows for structured composition: verbs and other relational words act as operations that combine or transform arguments. For instance, in the sentence “I like quantum,” the verb “like” takes two arguments (“I” and “quantum”) and is modeled as a unitary operation acting on their corresponding quantum states. This compositional encoding enables the model to reflect grammatical structure and semantic interactions in a mathematically grounded way, which is particularly well-suited for implementation as a quantum circuit.

\begin{figure}[h!]
    \centering
    \begin{minipage}{0.85\columnwidth}
        \centering
        \includegraphics[width = \columnwidth]{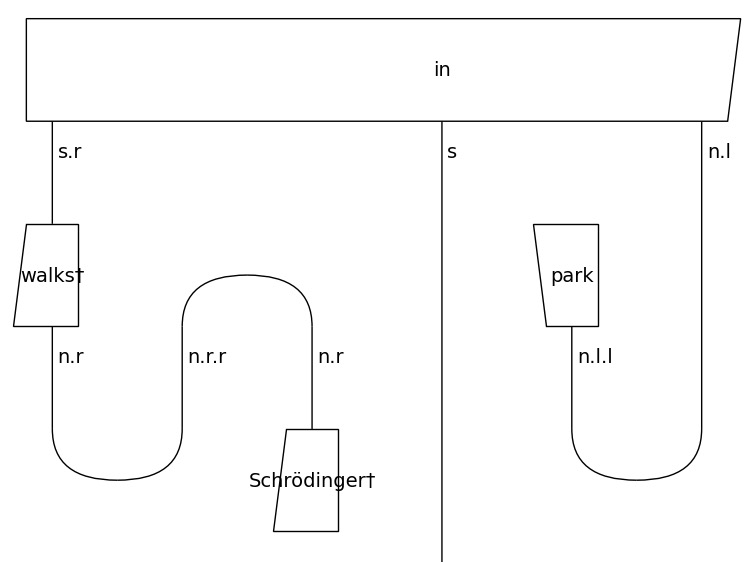}
    \caption{\footnotesize{Diagram representing “Schr\"odinger walks in the park.”}}
    \label{fig:sentence-diagram}
    \end{minipage}
\end{figure}

The diagram in Figure~\ref{fig:sentence-diagram} illustrates the DisCoCat quantum circuit representation of the sentence “Schr\"odinger walks in the park.” This diagram shows how a sentence can be composed diagrammatically from its parts using quantum circuits. The nouns “Schr\"odinger” and “park” serve as inputs—each assigned a noun qubit. The verb “walks” acts as an operator on “Schr\"odinger”, transforming the noun state and initiating a sentence-level meaning. Next, the preposition “in” combines the partial meaning “Schr\"odinger walks” with “park”, producing a complete sentence representation.

Formally, the diagram in Figure~\ref{fig:sentence-diagram} distinguishes between atomic types, such as noun and sentence, by assigning each a defined set of qubits in the circuit. Content words like “walks” and “in” are implemented as unitary operations that act on and often entangle these qubits to encode semantic composition. For example, “walks” consumes a noun input and produces a sentence output via a controlled gate, entangling the corresponding qubits. The final meaning of the sentence is encoded in the amplitude of the sentence qubit, which can be interpreted algebraically or approximated through quantum measurement. This quantum state serves as a compact, compositional representation of meaning and can be compared to other sentence states for tasks like semantic inference.

\begin{figure}[h!]
    \centering
    \begin{minipage}{\columnwidth}
        \centering
        \includegraphics[width = \columnwidth]{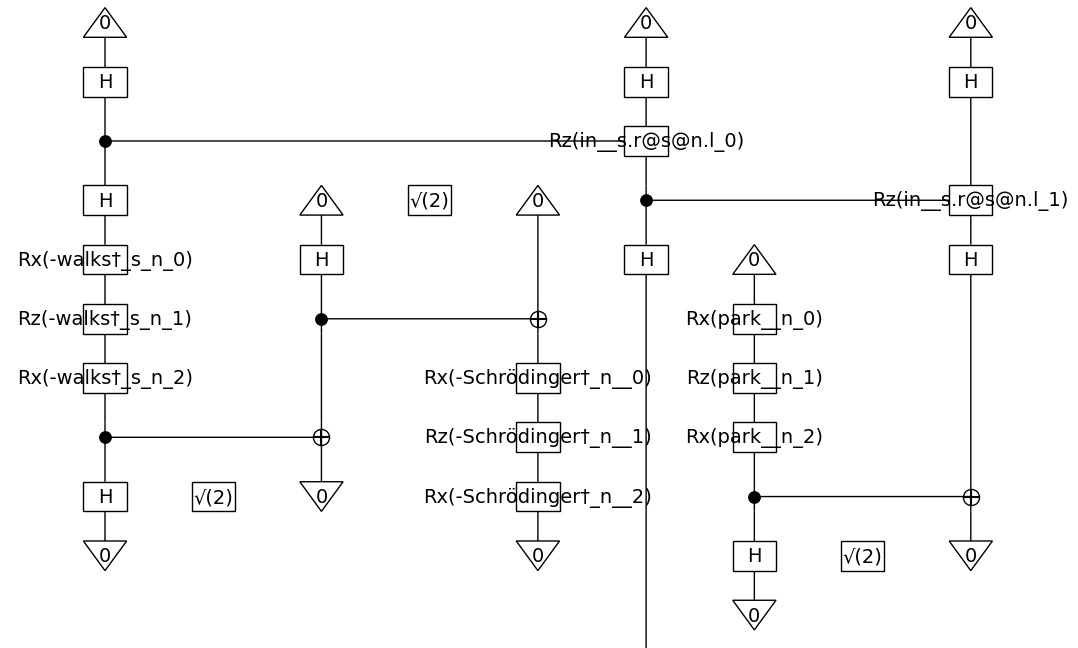}
    \caption{\footnotesize{Quantum circuit representing “Schr\"odinger walks in the park.”}}
    \label{fig:sentence-circuit}
    \end{minipage}
\end{figure}

To construct such circuits, we define an ansatz—a template that determines the circuit’s structure and expressiveness. The ansatz specifies how word-level quantum subsystems are initialized and how interactions (e.g., entangling gates) are applied to model compositional semantics. Each word in the sentence is mapped to one or more qubits and encoded using parameterized rotation and entangling gates. These parameters are optimized during training. The result is a parameterized quantum circuit in which each gate corresponds to a linguistic transformation, and the sentence meaning emerges as a distributed quantum state optimized for inference, as in Figure~\ref{fig:sentence-circuit}. Due to the exponential scaling of classical simulation, we limit our implementation to one qubit per word, which balances representational capacity with tractability. 

\section{Experiments} \label{experiment}
Our experiments investigate two core tasks: \textit{semantic relatedness} and \textit{natural language inference} (see Section~\ref{dataset}), evaluated across three categories of models—quantum, hybrid, and classical (Section~\ref{models}). The objective is to assess each model’s ability to learn compositional meaning and generalize under limited supervision.

\subsection{Dataset}\label{dataset}
We use the \textbf{Sentences Involving Compositional Knowledge (SICK)} dataset \citep{marelli-etal-2014-sick}, a widely adopted benchmark in natural language inference (NLI), including by \citet{hu2019monalog}.\footnote{\url{https://huggingface.co/datasets/RobZamp/sick}} The full dataset contains approximately 9,840 English sentence pairs, derived from existing corpora. For this study, we extract a subset of 100 sentence pairs, filtering to ensure each sentence contains no more than 11 words. This constraint reflects the computational limits of classical simulation of quantum circuits, while also creating a genuine few-shot learning scenario. Our train–dev–test split is set to 70:15:15.

The SICK dataset includes human-annotated labels for:
\paragraph{Semantic relatedness:} A continuous score from 1 to 5, which we normalize to the [0, 1] range.
\paragraph{Inference labels:} Categorical values indicating whether the hypothesis entails, contradicts, or is neutral with respect to the premise. Since entailment is not a symmetric relation, we evaluate both directions of each sentence pair. For instance, if sentence A entails sentence B but not the reverse, we label the pair \textit{entailment}, and add a second instance with the order reversed, labeled \textit{neutral}. This bidirectional expansion encourages the model to learn the asymmetry of inference relations.

\subsection{Metrics}
\subsubsection{Performance Evaluation Metrics}

We use different evaluation metrics for the two tasks in our benchmark to better reflect their distinct objectives:

\paragraph{Mean Squared Error (MSE)} is used for the \textit{semantic relatedness} task, as it involves predicting a continuous similarity score. This metric enables a unified comparison across models by measuring how closely predicted scores match the normalized human ratings.
    
\paragraph{Macro-averaged F1 score} is used for the \textit{inference classification} task, where the goal is to predict categorical labels (entailment, neutral, contradiction). Macro F1 provides a class-balanced view of performance by averaging F1 scores across all classes, regardless of label frequency.

As a result, our evaluation combines regression-based and classification-based metrics to provide a fuller picture of model behavior across tasks.

\subsubsection{Information-Theoretic Efficiency Metrics}

To quantify how much useful information a model gains per parameter update, we define the Information Gain per Parameter (IGPP) and its gradient-normalized variant (IGGP) as follows:

\begin{flalign}
&I(Y;\hat Y) = \sum_{y,\hat y} p(y,\hat y)\,\log\!\frac{p(y,\hat y)}{p(y)\,p(\hat y)} \\
&IGPP_t = \frac{\Delta I_t}{P} \\
&IGGP_t = \frac{\Delta I_t}{P\,\bigl\|\nabla_\theta L_t\bigr\|_2}\,
\end{flalign}

Here, \(I_t\) is the mutual information between predicted and true labels at epoch \(t\), \(P\) is the number of trainable parameters, and \(\|\nabla_\theta L_t\|\) is the L2 norm of model gradients. This allows fair comparison between quantum (qubit-limited) and classical (parameter-rich) models.

\subsection{Model Architecture}\label{models}

This experiment evaluates three types of models: classical transformer model, hybrid quantum-classical model, and pure quantum model

\subsubsection{Transformer Baseline (SBERT)}
\label{sec:sbert-model}

As a classical baseline, we employ the \textbf{Sentence-BERT (SBERT)} architecture, which modifies BERT for sentence-pair tasks by adding a pooling operation to produce fixed-length sentence embeddings \citep{reimers2019sentence}. We evaluate two variants of this model:

\begin{itemize}
    \item \textbf{SBERT-pretrained:} A MiniLM-L6-v2 model pretrained on large-scale sentence similarity datasets, fine-tuned on our relatedness and inference tasks.
    \item \textbf{SBERT-random:} The same architecture with randomly initialized weights and no pretraining, trained from scratch on our 100-example dataset.
\end{itemize}

Both versions take sentence pairs as input, encode them using a transformer encoder, and extract pooled embeddings from the [CLS] token. These embeddings are passed through a lightweight feedforward classification or regression head, depending on the task. The same architecture is used for both semantic relatedness (regression) and inference (classification).

While the pretrained model offers a strong upper bound for small-scale NLI tasks, the random model serves a different purpose: it acts as a fair baseline for comparison with quantum models, which lack pretraining and operate under similarly low-resource constraints. This setup allows us to assess not only absolute task performance but also learning efficiency and generalization in the few-shot regime.

\subsubsection{Hybrid Quantum-Classical Model (XOR Pipeline)}
\label{sec:hybrid-xor}

The hybrid model integrates a compositional quantum circuit with a lightweight classical neural network, enabling joint learning over symbolic structure and data-driven inference. Each sentence is parsed using a categorical grammar and converted into a parameterized quantum circuit via the lambeq-to-PennyLane pipeline. The circuit encodes sentence semantics based on the DisCoCat framework, where each word corresponds to a subcircuit defined by a chosen ansatz (e.g., IQPAnsatz). Atomic types such as \textit{Noun} and \textit{Sentence} are mapped to dedicated qubits, and functional words (e.g., verbs, prepositions) act as entangling gates that combine their inputs.

The quantum circuits are executed in simulation using PennyLane, with expectation values measured at designated wires. These measurements form a fixed-length quantum embedding for the input sentence or sentence pair. This embedding is passed to a classical feedforward network, XORNet, which maps the quantum outputs to a scalar prediction. The network consists of one or two dense layers with non-linear activations and is trained jointly with the quantum parameters via the Adam optimizer.

This hybrid architecture allows the quantum circuit to serve as a semantic feature extractor—capturing structural and entangled representations—while the classical component introduces non-linearity and flexibility. The model is used for both regression (relatedness) and classification (inference) tasks, with a unified architecture across tasks.

\subsubsection{Pure Quantum Model (KL Divergence)}
\label{sec:quantum-kl}

The KL model is a fully quantum, compositional architecture designed to perform the NLI task directly, with no classical post-processing. Following the framework proposed by \citet{sadrzadeh2018sentence}, the model interprets entailment as a containment relation between quantum representations of sentence meaning. Each sentence is first parsed into a string diagram using lambeq and compiled into a parameterized quantum circuit via an ansatz such as IQPAnsatz. Words are mapped to sequences of quantum gates, and their syntactic roles determine how they act on sentence-level or noun-level qubits. Functional words like verbs and prepositions are implemented as entangling operations, enabling the circuit to encode compositional semantics.

Once compiled and simulated using PennyLane, the circuit produces a quantum state \(\rho\) for each sentence. Given a sentence pair \((s_1, s_2)\), their quantum states \(\rho_1\) and \(\rho_2\) are interpreted as probabilistic encodings over measurement outcomes (i.e., bitstring distributions). The degree to which \(\rho_1\) entails \(\rho_2\) is then operationalized by computing the Kullback–Leibler (KL) divergence between the two distributions:
\[
D_{{KL}}(P_{\rho_2} \| P_{\rho_1}) = \sum_i P_{\rho_2}(i) \log \frac{P_{\rho_2}(i)}{P_{\rho_1}(i)}
\]
A lower KL divergence indicates that \(\rho_1\) captures most of the probability mass of \(\rho_2\), thus supporting an entailment relation.

Training is conducted by minimizing a cross-entropy loss between predicted and target labels, with the KL divergence serving as the core similarity signal. While originally designed for inference, we extend this model to support both classification (entailment) and regression (relatedness) tasks by mapping the divergence output to either categorical labels or continuous similarity scores. The model remains fully quantum—without classical neural components—providing a semantically grounded and interpretable approach to meaning comparison directly through quantum state relationships.

\section{Results}\label{results}

Our results demonstrate that quantum-based models achieve comparable performance to classical baselines on both semantic relatedness and inference tasks (see Section~\ref{performance}). Despite having significantly fewer parameters, the quantum and hybrid models exhibit strong learning capabilities. Notably, they achieve orders-of-magnitude greater efficiency, as measured by our proposed metric—Information Gain per Parameter (IGPP)—with some models propagating up to $10^5$ times more information per parameter during training compared to classical transformers (see Section~\ref{efficiency}). However, the training dynamics reveal that quantum circuit models struggle with generalization, largely due to circuit isolation—the lack of shared parameters across input circuits (see Section~\ref{train}).

\subsection{Model performance} \label{performance}

Table~\ref{tab:full-comparison} summarizes the performance of all models across the two tasks. 

On the semantic relatedness task, the Q-XOR model outperforms all others, including pretrained SBERT, with the lowest test MSE of $0.009$. This suggests that quantum circuits can capture fine-grained semantic similarity effectively, especially in low-resource settings. Interestingly, even SBERT with random initialization performs well on relatedness (MSE = $0.017$), indicating that the task may require less compositional structure than inference.

The inference task, however, proves more challenging and reveals a clearer separation in generalization ability. As expected, the pretrained SBERT model performs best on the inference task, achieving a macro F1 score of 0.690. However, this comes at the cost of massive scale, with over 100 million trainable parameters. While both Q-XOR and Q-KL match the random SBERT baseline (F1 = $0.232$), they fail to surpass it, highlighting the difficulty of achieving robust generalization using circuit-isolated quantum models. These results suggest that while quantum models are highly efficient and promising for tasks requiring localized semantic encoding, their generalization remains limited without mechanisms for broader structural reuse or parameter sharing.

\begin{table}[t]
\centering
\caption{Model size (trainable DoF), training epochs, and test MSE for Relatedness and test macro F1 for Inference.}
\begin{tabular}{llccc}
\hline
\textbf{Model} & \textbf{Embed} & \textbf{DoF} 
& \textbf{Relatedness} 
& \textbf{Inference} \\
\hline
SBERT           & Pretrained    & $\sim$109M & 0.014 & \textbf{0.690} \\
SBERT           & Random        & $\sim$109M & 0.017 & 0.232 \\
Q-XOR        & N/A        & 1,042      & \textbf{0.009} & 0.232 \\
Q-KL         & N/A        & 981        & 0.038 & 0.232 \\
Q-Cluster    & N/A        & 799        & 0.012 & 0.467 \\
\hline
\end{tabular}
\label{tab:full-comparison}
\end{table}

\subsection{Information Gain during Training} \label{efficiency}
Table~\ref{IGGP} presents the peak Information Gain per Parameter (IGPP) achieved by each model during training. The results highlight a stark contrast in learning efficiency between quantum and classical architectures. The Quantum XOR model achieves the highest IGPP value, indicating that it extracts the most information per trainable parameter per epoch. It is closely followed by the Quantum KL model, which also demonstrates strong efficiency.

In comparison, the SBERT model with random embeddings performs significantly worse, achieving an IGPP score that is four to five orders of magnitude lower. Despite having over 100 million parameters, the classical model exhibits early saturation and contributes negligible incremental information after initial epochs. This suggests that quantum architectures are vastly more efficient at propagating and encoding useful information per parameter, particularly in low-resource settings. The high IGPP values observed in the quantum models support the hypothesis that entangled quantum representations can capture structured semantic relationships more compactly and efficiently than classical models. These findings emphasize the promise of quantum NLP models in tasks requiring semantic generalization under parameter and data constraints.

\begin{table}[h]
\centering
\caption{Peak information gain per parameter (\(\max_t IGPP_t\)) across models. Quantum models show orders of magnitude higher parameter efficiency compared to the classical baseline.} \label{IGGP}
\begin{tabular}{lc}
\hline
\textbf{Model} & \(\boldsymbol{\max_t IGPP_t}\) \\
\hline
Classical SBERT (random embeddings) & \(3 \times 10^{-9}\)   \\
Quantum XOR circuit        & \(\mathbf{1.5 \times 10^{-4}}\) \\
Quantum KL model           & \(9 \times 10^{-5}\)   \\
\hline
\end{tabular}
\end{table}

Appendix Figure~\ref{fig:igpp-dynamics} presents the epoch-wise evolution of two key metrics—IGPP and IGGP—which quantify learning efficiency per parameter during training on the inference task. These curves provide insight into how effectively different models propagate information, independent of their final accuracy. As expected, the SBERT model with random initialization exhibits consistently low IGPP values, fluctuating around $10^{-9}$, indicating weak learning dynamics in the few-shot regime. Most of its high-capacity parameter space remains underutilized when trained from scratch on limited data.

In contrast, the quantum models show substantially higher and more dynamic IGPP and IGGP values, reflecting stronger parameter updates and more efficient learning. The Quantum-XOR model achieves the sharpest peaks, in line with its highest IGPP score reported in Table~\ref{IGGP}, though it also displays greater volatility. The Quantum-KL model shows comparatively smoother curves, suggesting more stable information flow throughout training. These trends reinforce the notion that quantum models, despite their compact parameter count, are highly information-efficient. Moreover, the correlation between IGPP and IGGP indicates that the observed gains arise not merely from gradient magnitude, but from structurally meaningful updates enabled by quantum entanglement and compositional architecture.

These efficiency trends are further supported by model selection criteria based on information theory. As shown in Table~\ref{tab:model_comparison} and Table~\ref{tab:classification_model_comparison}, quantum models—despite their drastically smaller parameter counts—achieve significantly lower Akaike Information Criterion (AIC) and Bayesian Information Criterion (BIC) scores compared to classical baselines. For the relatedness task, Quantum-XOR achieves a validation MSE of $0.0094$ and an AIC of just $1.90e+03$, whereas both SBERT variants yield AICs exceeding $2.19e+08$—highlighting the inefficiency of over-parameterized models in few-shot settings. Similarly, for the inference classification task, Quantum-KL produces the lowest AIC ($2.19e+03$) and BIC ($4.77e+03$) among all evaluated models, despite not having the best cross-entropy score. These findings reinforce the conclusion that quantum models offer a superior trade-off between fit and complexity, with architectures like Quantum-KL and XOR able to extract semantically meaningful representations efficiently—even under tight data and parameter constraints. This supports the hypothesis that entangled quantum encodings not only propagate more meaningful gradients but also result in structurally leaner, more generalizable models.

\begin{table}[t]
\centering
\captionsetup{width=\columnwidth}
\caption{Validation MSE, log-likelihood (LogL), AIC, and BIC. Lower AIC/BIC = better fit-complexity trade-off.}
\begin{tabular}{lcccc}
\hline
\textbf{Model} & \textbf{MSE} & \textbf{LogL} & \textbf{AIC} & \textbf{BIC} \\
\hline
SBERT-random        & 0.0168 & 62.5 & $2.19e+08$ & $5.05e+08$ \\
SBERT-pretrained    & 0.0145 & 69.8 & $2.19e+08$ & $5.05e+08$ \\
Quantum-XOR         & 0.0094 & 91.5 & $1.90e+03$ & $4.62e+03$ \\
Quantum-KL          & 0.0382 & 21.4 & $1.92e+03$ & $4.47e+03$ \\
\hline
\end{tabular}
\label{tab:model_comparison}
\end{table}

\begin{table}[t]
\centering
\captionsetup{width=\columnwidth}
\caption{Test Cross-Entropy (CE), log-likelihood (LogL), AIC, and BIC for classification models. Lower AIC/BIC = better fit-complexity trade-off.}
\begin{tabular}{lcccc}
\hline
\textbf{Model} & \textbf{CE} & \textbf{LogL} & \textbf{AIC} & \textbf{BIC} \\
\hline
SBERT-random        & 1.1651 & -116.5 & $2.19e+08$ & $5.05e+08$ \\
SBERT-pretrained    & 0.6648 & -66.5  & $2.19e+08$ & $5.05e+08$ \\
Quantum-XOR         & 1.1446 & -114.5 & $2.36e+03$ & $5.13e+03$ \\
Quantum-KL          & 1.0091 & -100.9 & $\mathbf{2.19e+03}$ & $\mathbf{4.77e+03}$ \\
\hline
\end{tabular}
\label{tab:classification_model_comparison}
\end{table}

\subsection{Training Dynamics} \label{train}
The training dynamics for both semantic relatedness and inference tasks reveal notable differences across models, particularly in terms of convergence stability and generalization.

As shown in Appendix Figure~\ref{fig:related-train-comparison}, all models perform relatively well on the relatedness task, with validation loss remaining low and stable throughout training. Notably, the Q-XOR model exhibits the fastest and smoothest convergence, with both training and validation loss decreasing consistently. SBERT (random) achieves stable, low validation loss despite a relatively high training loss, suggesting that its pretrained embeddings—even when unfrozen—provide useful inductive biases in few-shot settings.

In contrast, the inference task (Appendix Figure~\ref{fig:inference-train-comparison}) presents a greater challenge. All models show more fluctuation in training and validation loss, particularly the quantum circuits. Both Q-KL and Q-XOR exhibit noisy and non-monotonic training behavior, highlighting the sensitivity of quantum models to gradient instability, circuit-specific variance, and the structural demands of NLI tasks. SBERT (random) fails to show meaningful convergence, with both training and validation losses remaining flat and high throughout.

These trends support our central observation: quantum models, while highly parameter-efficient and capable of capturing structured semantics, are more vulnerable to instability and generalization limits in low-data regimes—especially when no parameter sharing or structural smoothing is applied.

\section{Cluster-based Parameter Sharing}\label{cluster}

We introduce a novel \textbf{cluster-based quantum architecture} designed to address a core limitation of existing QNLP models: \textit{circuit isolation}. In conventional quantum NLP pipelines, each sentence is compiled into a unique quantum circuit, with no parameter sharing between inputs. This prevents generalization across structurally similar examples and severely restricts learning efficiency. Our model tackles this problem by introducing a mechanism for parameter sharing through \textbf{word-level clustering}, allowing related words to share circuit parameter distributions during training.

\subsection{Proposed Cluster-Based Quantum Model}
\label{sec:cluster-model}

The architecture begins by extracting contextual word embeddings using a classical language model—specifically, a pretrained SBERT encoder. Each word is also assigned a syntactic type via DisCoCat string diagrams, ensuring structural consistency with the Categorical Compositional Distributional framework. We then apply a K-nearest neighbors (KNN) algorithm to cluster words based on their SBERT embeddings and syntactic roles. This creates a mapping from words to shared \textit{semantic clusters}, serving as a computationally efficient way to leverage large classical LLM representations without directly translating dense embeddings to quantum gates, which remains resource-prohibitive on current quantum hardware.

Each sentence is then converted into a quantum circuit, where word-level subcircuits reference their assigned clusters rather than using unique parameter sets. Gate parameters within the circuit are no longer trainable directly, but are \textbf{sampled from learned distributions}—specifically, a Gaussian parameterized by cluster-level mean \(\mu_c\) and variance \(\rho_c\). This design enables circuits to remain fully compositional while promoting \textit{parameter sharing} across structurally or semantically similar words.

The circuit output, as in previous quantum models, is a 2-dimensional amplitude vector obtained via measurement. This output is passed to a downstream \textbf{order embedding layer}, which projects the state into a partially ordered latent space. The embedding is then mapped to a scalar prediction for the semantic relatedness task or a 3-dimensional output for classification (entailment, contradiction, neutral). Importantly, the only trainable parameters in the model are the cluster-level distribution parameters (\(\mu, \rho\)) and the weights of the order embedding layer.

This design offers three major advantages: (1) it enables \textbf{parameter efficiency} by reducing redundancy across sentence-specific circuits; (2) it introduces \textbf{generalization through homogeneity}, allowing unseen combinations of familiar elements to be composed from shared latent spaces; and (3) it provides a scalable way to incorporate classical semantic structure into quantum models without incurring the cost of direct quantum translation of large embedding vectors.

\subsection{Cluster Model Performance and Efficiency}
\label{sec:cluster-results}

\begin{table}[t]
\centering
\captionsetup{width=\columnwidth}
\caption{Summary of Q-Cluster model performance, with best value among other models (SBERT-random, Q-XOR, Q-KL) for reference.}
\begin{tabular}{lcc}
\hline
\textbf{Metric} & \textbf{Q-Cluster} & \textbf{Best (Others)} \\
\hline
\multicolumn{3}{l}{\textbf{Model Size}} \\
DoF & 799 & 981 (Q-KL) \\
\hline
\multicolumn{3}{l}{\textbf{Task Performance}} \\
MSE (Rel) & 0.012 & \textbf{0.009} (Q-XOR) \\
F1 (Inf) & \textbf{0.467} & 0.232 (Q-XOR / Q-KL) \\
CE (Inf) & 1.10 & \textbf{1.01} (Q-KL) \\
\hline
\multicolumn{3}{l}{\textbf{Information Efficiency}} \\
IGPP$_\text{max}$ & \(8\!\times\!10^{-5}\) & \textbf{\(1.5\!\times\!10^{-4}\)} (Q-XOR) \\
\hline
\multicolumn{3}{l}{\textbf{Model Fit (Relatedness)}} \\
LogL & 78.4 & \textbf{91.5} (Q-XOR) \\
AIC & \textbf{1.44e3} & 1.90e3 (Q-XOR) \\
BIC & \textbf{3.52e3} & 4.62e3 (Q-XOR) \\
\hline
\multicolumn{3}{l}{\textbf{Model Fit (Inference)}} \\
LogL & -110.2 & \textbf{-100.9} (Q-KL) \\
AIC & \textbf{1.82e3} & 2.19e3 (Q-KL) \\
BIC & \textbf{3.90e3} & 4.77e3 (Q-KL) \\
\hline
\end{tabular}
\label{tab:cluster-summary}
\end{table}


\paragraph{Task Performance.}
As shown in Table~\ref{tab:cluster-summary}, the Quantum-Cluster model achieves a macro F1 score of \textbf{0.467} on the inference task— substantially outperforming other quantum variants (Q-XOR and Q-KL, both at 0.232) and narrowing the gap with the pretrained SBERT baseline (0.690). On the semantic relatedness task, it also performs well, with a test MSE of 0.012—competitive with SBERT and outperforming the Q-KL model. These results demonstrate that parameter sharing via clustering significantly improves task-level generalization in low-resource quantum settings.

\paragraph{Training Dynamics.}
The training curves in Appendix Figure~\ref{fig:related-train-comparison} and~\ref{fig:inference-train-comparison} reveal that the Quantum-Cluster model exhibits moderately noisy convergence. While its training loss decreases consistently, the validation loss on the inference task shows non-monotonic fluctuations, especially in early epochs. This instability is expected in few-shot scenarios and may reflect inter-cluster variability or early overfitting before convergence stabilizes. Still, the final validation loss remains lower than in the Q-KL model, and the relatedness task converges stably.

\paragraph{Information Efficiency.}
Efficiency metrics in Appendix Figure~\ref{fig:igpp-dynamics} highlight the strength of the Cluster model. It achieves a peak IGPP of $8 \times 10^{-5}$, and its epoch-wise IGPP and IGGP curves are relatively smooth compared to Q-XOR, indicating more stable learning. These dynamics suggest that parameter sharing through clustering not only improves generalization but also leads to more consistent information flow during training. The cluster model is the most balanced of the quantum architectures, combining improved inference performance with high training efficiency and moderate gradient stability.

Taken together, these results show that the Cluster model provides a promising strategy for overcoming structural limitations in quantum NLP pipelines, offering better generalization without sacrificing the representational power of compositional quantum circuits.

\section{Discussion}\label{discussion}
To further assess the generalization abilities of our models beyond standard evaluation metrics, we conduct targeted probing on two key linguistic dimensions: lexical entailment and compositional inference. Table~\ref{tab:word-subset} presents predictions on word and phrase pairs involving subset or hyponymic relations, such as “children” and “kids”, designed to test whether models capture hierarchical semantic relationships at the lexical level. Following this, Table~\ref{tab:compositional} examines model behavior on sentence pairs constructed to isolate compositional phenomena—including negation, adverbial modification, and argument structure. These tasks go beyond surface-level similarity, requiring models to reason about meaning in structured and often asymmetric ways. Together, they provide insight into how well quantum models internalize linguistic structure and whether architectural innovations like clustering improve generalization across different levels of language.

\paragraph{Word-level relations}
Table~\ref{tab:word-subset} examines model predictions on word and phrase pairs involving lexical entailment and subset relations, such as hypernyms (e.g., ``children'' is a hypernym of ``boys'') and hyponyms (e.g., ``tree'' is a hyponyms of ``plant''). The gold labels reflect hyponyms (2) or hypernyms (1), corresponding to entailment (2) and neutral (1) in an upward entailment environment. The results show that most models fail to capture these fine-grained lexical relations. SBERT and XOR consistently predict hypernym (1) across nearly all examples, suggesting that without explicit lexical training or external knowledge sources, they struggle to infer subset relationships from few-shot data. The KL model, however, displays slightly more variability: it correctly identifies some hyponym pairs (e.g., ``little boys'' $\rightarrow$ ``kids'') and appears more sensitive to hierarchical lexical structures. Still, it also overgeneralizes in several cases, labeling hypernym relations (1) like ``children'' $\rightarrow$ ``boys'' or ``kids'' $\rightarrow$ ``boys'' as hyponyms (2). These findings reinforce a key limitation: current quantum models lack robust lexical generalization, particularly in the absence of large-scale pretraining. While the KL model shows the most promising lexical behavior, it remains constrained by data sparsity and the absence of structured lexical priors.

\begin{table}[h]
\centering
\caption{Predictions for word-level subset relations.}
\begin{tabular}{llccccc}
\hline
\textbf{A} & \textbf{B} & \textbf{G} & \textbf{BERT} & \textbf{Xor} & \textbf{KL} & \textbf{Cluster} \\
\hline
children & kids & 2 & 1 & 1 & 2 & 1\\
children & boys & 1& 1& 1 & 2 & 1\\
kids & boys & 1 &1 & 1 & 2 & 1\\
young boys & little boys & 2 & 1& 1 & 2 & 1\\
young boys & boys & 2 & 1 & 1 & 1 & 1\\
little boys & boys & 2 & 1 & 1 & 2 & 1\\
little boys & kids & 2 & 1 & 1 & 2 & 1\\
tree & plant & 2 & 2 & 1 & 1 & 1\\
\hline
\end{tabular}
\label{tab:word-subset}
\end{table}

\paragraph{Compositional patterns}
Table~\ref{tab:compositional} presents model predictions on controlled sentence pairs designed to test compositional inference abilities—specifically, how well models handle logical relationships involving negation, adverbial modification, and semantic containment. The ground-truth labels reflect entailment (2), contradiction (0), or neutrality (1). Overall, the results reveal clear differences in compositional generalization. SBERT and the XOR model consistently predict neutral scores (1), suggesting limited sensitivity to fine-grained compositional shifts. The KL model frequently overpredicts entailment (label 2), particularly in cases involving negation (e.g., ``fence is playing'' vs. ``fence isn't playing''), indicating that while it detects some structural asymmetry, it struggles with semantic nuance. Notably, the Cluster model provides a slightly more diverse response pattern, correctly predicting a neutral label in the entailment-reducing pair (``fence is riding the tree dangerously'' vs. ``fence is riding the tree''). However, its predictions still tend to gravitate toward neutrality, revealing an under-generalization bias. These results suggest that while none of the models demonstrate strong compositional reasoning, the Cluster model shows emerging sensitivity to structural differences—highlighting the promise of parameter sharing as a path toward improved generalization in quantum NLI systems.

\begin{table}[h!]
\centering
\caption{Predictions for compositional sentence-level relations.}
\begin{tabular}{llccccc}
\hline
\textbf{A} & \textbf{B} & \textbf{Gold} & \textbf{SBERT} & \textbf{XOR}  & \textbf{KL}  & \textbf{Cluster} \\
\hline
fence is riding the tree & fence isn't riding the tree &0 & 1 & 1 & 2 & 1\\
fence is playing & fence isn't playing &0 & 1 & 1 & 2 & 2 \\
fence is playing dangerously & fence isn't playing dangerously &0 & 1 & 1 & 2 & 1 \\
fence isn't playing dangerously & fence isn't playing & 1 & 1 & 1 & 1 & 1\\
fence isn't playing & fence isn't playing dangerously & 2 & 1 & 1 & 1 & 1 \\
\hline
\end{tabular}
\label{tab:compositional}
\end{table}

\paragraph{Data Homogeneity}
A persistent challenge in our experiments is the lack of data homogeneity, particularly in the quantum setting. Because each sentence is compiled into a structurally unique circuit, there is little to no parameter sharing across inputs. This limits the model’s ability to generalize: training on one sentence offers no direct benefit for unseen ones, even when they are semantically similar. The result is highly variable validation performance despite steady training loss improvements. Our proposed cluster model addresses this partially by promoting shared circuit parameter distributions. Further work is needed to enable more robust parameter reuse across structurally or lexically similar inputs.

\paragraph{Simulation Bottlenecks}
Quantum circuit simulation remains a major computational bottleneck. Even with modest sentence lengths and limited qubit allocations, classical simulation incurs exponential time and memory costs due to the tensor-product structure of quantum states. This restricts model scale and batch throughput during training. However, on actual quantum hardware, these operations would be performed in parallel via quantum superposition, with time complexity driven primarily by gate depth and width—often polynomial for practical ansätze. Real hardware deployment is thus key to unlocking the true scalability of QNLP models.

\section{Conclusion}\label{conclusion}

This work presents a comprehensive study of quantum and hybrid models for Natural Language Inference (NLI) under realistic constraints—limited data and qubit resources. We introduced a few-shot benchmark using only 100 training examples, enabling fair comparisons between classical and quantum approaches. Across this setting, our experiments demonstrate that quantum models can achieve performance comparable to classical transformers while operating with orders-of-magnitude fewer parameters. Through our proposed metric, \textit{Information Gain per Parameter} (IGPP), we show that quantum models are significantly more efficient in propagating learning signal, reflecting their potential for high structural expressiveness even in low-data regimes.

We further identify a critical bottleneck in existing quantum NLP models: \textit{circuit isolation}, which limits generalization by preventing parameter sharing across inputs. To address this, we propose a novel cluster-based quantum model that introduces shared parameter distributions over clustered word embeddings. Our Cluster model achieves the best macro F1 among all quantum models and demonstrates improved generalization while preserving compositional structure. Importantly, it provides a scalable pathway for integrating classical semantic information into quantum circuits without incurring prohibitive translation costs.

Taken together, these results highlight the promise of quantum models—especially when compositional and parameter-sharing strategies are explicitly designed—to perform structured language tasks efficiently. Our findings suggest that with continued refinement, quantum NLP can offer not just theoretical novelty but also practical value in low-resource, structure-sensitive settings. Future work will focus on extending these methods to richer datasets and deploying models on real quantum hardware to explore their true computational advantages.

\section{Limitations and Future Work}\label{limitation}

While our results highlight the promise of quantum and hybrid models for semantic inference, several limitations remain—both practical and methodological.

\paragraph{Quantum Hardware Deployment}
All experiments in this work are performed via classical simulation, which imposes strict limits on circuit size and batch throughput. Realizing these models on actual quantum hardware is a critical next step to validate their efficiency claims and unlock their potential for scalability.

\paragraph{Generalization and Circuit Isolation}
Although our proposed cluster-based model improves generalization by introducing parameter sharing, circuit-level isolation remains a structural bottleneck in quantum NLP. Future work will explore complementary strategies such as \textit{circuit stratification}—pre-grouping inputs with similar grammatical or semantic structure to promote shared computation paths.

\paragraph{Parameter Sharing and Structural Priors}
Our findings suggest that shared representations (e.g., cluster-level parameter distributions) are essential for learning generalizable patterns from few-shot data. Further architectural innovations—such as hierarchical parameter templates or dynamically constructed ansätze—may improve the model’s ability to internalize linguistic structure.

Overall, future research will focus on integrating symbolic priors, scaling circuit design, and leveraging quantum hardware to extend the practical viability of QNLP systems in real-world language understanding tasks.

\section{Acknowledgement}
This work was developed as part of a course project for a class taught by Dr. Mohsen Heidari. The authors would like to thank him for his valuable advice and insightful suggestions, which were instrumental in shaping the direction of this project.
\section{Declaration on the Use of AI}
This paper includes content generated with the assistance of AI tools, primarily for language refinement and code formatting. All model design, experimental work, data analysis, and interpretative conclusions were carried out and verified solely by the authors.

\bibliographystyle{ACM-Reference-Format}
\bibliography{references}


\begin{thebibliography}{15}


\ifx \showCODEN    \undefined \def \showCODEN     #1{\unskip}     \fi
\ifx \showDOI      \undefined \def \showDOI       #1{#1}\fi
\ifx \showISBNx    \undefined \def \showISBNx     #1{\unskip}     \fi
\ifx \showISBNxiii \undefined \def \showISBNxiii  #1{\unskip}     \fi
\ifx \showISSN     \undefined \def \showISSN      #1{\unskip}     \fi
\ifx \showLCCN     \undefined \def \showLCCN      #1{\unskip}     \fi
\ifx \shownote     \undefined \def \shownote      #1{#1}          \fi
\ifx \showarticletitle \undefined \def \showarticletitle #1{#1}   \fi
\ifx \showURL      \undefined \def \showURL       {\relax}        \fi
\providecommand\bibfield[2]{#2}
\providecommand\bibinfo[2]{#2}
\providecommand\natexlab[1]{#1}
\providecommand\showeprint[2][]{arXiv:#2}

\bibitem[Bernardi et~al\mbox{.}(2007)]%
        {bernardi2007lite}
\bibfield{author}{\bibinfo{person}{Raffaella Bernardi}, \bibinfo{person}{Diego Calvanese}, {and} \bibinfo{person}{Camilo Thorne}.} \bibinfo{year}{2007}\natexlab{}.
\newblock \showarticletitle{Lite natural language}. In \bibinfo{booktitle}{\emph{Proceedings of the 7th International Workshop on Computational Semantics (IWCS-7)}}, Vol.~\bibinfo{volume}{217}.
\newblock


\bibitem[Bradley et~al\mbox{.}(2018)]%
        {bradley2018translating}
\bibfield{author}{\bibinfo{person}{Tai-Danae Bradley}, \bibinfo{person}{Martha Lewis}, \bibinfo{person}{Jade Master}, {and} \bibinfo{person}{Brad Theilman}.} \bibinfo{year}{2018}\natexlab{}.
\newblock \showarticletitle{Translating and Evolving: Towards a model of language change in DisCoCat}.
\newblock \bibinfo{journal}{\emph{arXiv preprint arXiv:1811.11041}} (\bibinfo{year}{2018}).
\newblock


\bibitem[Chen et~al\mbox{.}(2021)]%
        {chen2021neurallog}
\bibfield{author}{\bibinfo{person}{Zeming Chen}, \bibinfo{person}{Qiyue Gao}, {and} \bibinfo{person}{Lawrence~S Moss}.} \bibinfo{year}{2021}\natexlab{}.
\newblock \showarticletitle{NeuralLog: Natural Language Inference with Joint Neural and Logical Reasoning}. In \bibinfo{booktitle}{\emph{Proceedings of* SEM 2021: The Tenth Joint Conference on Lexical and Computational Semantics}}. \bibinfo{pages}{78--88}.
\newblock


\bibitem[Coecke et~al\mbox{.}(2010)]%
        {coecke2010mathematical}
\bibfield{author}{\bibinfo{person}{Bob Coecke}, \bibinfo{person}{Mehrnoosh Sadrzadeh}, {and} \bibinfo{person}{Stephen Clark}.} \bibinfo{year}{2010}\natexlab{}.
\newblock \showarticletitle{Mathematical foundations for a compositional distributional model of meaning}.
\newblock \bibinfo{journal}{\emph{arXiv preprint arXiv:1003.4394}} (\bibinfo{year}{2010}).
\newblock


\bibitem[Demszky et~al\mbox{.}(2018)]%
        {demszky2018transforming}
\bibfield{author}{\bibinfo{person}{Dorottya Demszky}, \bibinfo{person}{Kelvin Guu}, {and} \bibinfo{person}{Percy Liang}.} \bibinfo{year}{2018}\natexlab{}.
\newblock \showarticletitle{Transforming question answering datasets into natural language inference datasets}.
\newblock \bibinfo{journal}{\emph{arXiv preprint arXiv:1809.02922}} (\bibinfo{year}{2018}).
\newblock


\bibitem[Dziri et~al\mbox{.}(2019)]%
        {dziri2019evaluating}
\bibfield{author}{\bibinfo{person}{Nouha Dziri}, \bibinfo{person}{Ehsan Kamalloo}, \bibinfo{person}{Kory~W Mathewson}, {and} \bibinfo{person}{Osmar Zaiane}.} \bibinfo{year}{2019}\natexlab{}.
\newblock \showarticletitle{Evaluating coherence in dialogue systems using entailment}.
\newblock \bibinfo{journal}{\emph{arXiv preprint arXiv:1904.03371}} (\bibinfo{year}{2019}).
\newblock


\bibitem[Hu et~al\mbox{.}(2019)]%
        {hu2019monalog}
\bibfield{author}{\bibinfo{person}{Hai Hu}, \bibinfo{person}{Qi Chen}, \bibinfo{person}{Kyle Richardson}, \bibinfo{person}{Atreyee Mukherjee}, \bibinfo{person}{Lawrence~S Moss}, {and} \bibinfo{person}{Sandra Kuebler}.} \bibinfo{year}{2019}\natexlab{}.
\newblock \showarticletitle{MonaLog: a lightweight system for natural language inference based on monotonicity}.
\newblock \bibinfo{journal}{\emph{arXiv preprint arXiv:1910.08772}} (\bibinfo{year}{2019}).
\newblock


\bibitem[MacCartney and Manning(2008)]%
        {maccartney-manning-2008-modeling}
\bibfield{author}{\bibinfo{person}{Bill MacCartney} {and} \bibinfo{person}{Christopher~D. Manning}.} \bibinfo{year}{2008}\natexlab{}.
\newblock \showarticletitle{Modeling Semantic Containment and Exclusion in Natural Language Inference}. In \bibinfo{booktitle}{\emph{Proceedings of the 22nd International Conference on Computational Linguistics (Coling 2008)}}, \bibfield{editor}{\bibinfo{person}{Donia Scott} {and} \bibinfo{person}{Hans Uszkoreit}} (Eds.). \bibinfo{publisher}{Coling 2008 Organizing Committee}, \bibinfo{address}{Manchester, UK}, \bibinfo{pages}{521--528}.
\newblock
\urldef\tempurl%
\url{https://aclanthology.org/C08-1066/}
\showURL{%
\tempurl}


\bibitem[Marelli et~al\mbox{.}(2014)]%
        {marelli-etal-2014-sick}
\bibfield{author}{\bibinfo{person}{Marco Marelli}, \bibinfo{person}{Stefano Menini}, \bibinfo{person}{Marco Baroni}, \bibinfo{person}{Luisa Bentivogli}, \bibinfo{person}{Raffaella Bernardi}, {and} \bibinfo{person}{Roberto Zamparelli}.} \bibinfo{year}{2014}\natexlab{}.
\newblock \showarticletitle{A {SICK} cure for the evaluation of compositional distributional semantic models}. In \bibinfo{booktitle}{\emph{Proceedings of the Ninth International Conference on Language Resources and Evaluation ({LREC}'14)}}. \bibinfo{publisher}{European Language Resources Association (ELRA)}, \bibinfo{address}{Reykjavik, Iceland}, \bibinfo{pages}{216--223}.
\newblock
\urldef\tempurl%
\url{http://www.lrec-conf.org/proceedings/lrec2014/pdf/363_Paper.pdf}
\showURL{%
\tempurl}


\bibitem[Miranda et~al\mbox{.}(2022)]%
        {miranda2022quantum}
\bibfield{author}{\bibinfo{person}{Eduardo~Reck Miranda}, \bibinfo{person}{Richie Yeung}, \bibinfo{person}{Anna Pearson}, \bibinfo{person}{Konstantinos Meichanetzidis}, {and} \bibinfo{person}{Bob Coecke}.} \bibinfo{year}{2022}\natexlab{}.
\newblock \showarticletitle{A quantum natural language processing approach to musical intelligence}.
\newblock In \bibinfo{booktitle}{\emph{Quantum Computer Music: Foundations, Methods and Advanced Concepts}}. \bibinfo{publisher}{Springer}, \bibinfo{pages}{313--356}.
\newblock


\bibitem[Reimers and Gurevych(2019)]%
        {reimers2019sentence}
\bibfield{author}{\bibinfo{person}{Nils Reimers} {and} \bibinfo{person}{Iryna Gurevych}.} \bibinfo{year}{2019}\natexlab{}.
\newblock \showarticletitle{Sentence-bert: Sentence embeddings using siamese bert-networks}.
\newblock \bibinfo{journal}{\emph{arXiv preprint arXiv:1908.10084}} (\bibinfo{year}{2019}).
\newblock


\bibitem[Richardson et~al\mbox{.}(2020)]%
        {richardson2020probing}
\bibfield{author}{\bibinfo{person}{Kyle Richardson}, \bibinfo{person}{Hai Hu}, \bibinfo{person}{Lawrence Moss}, {and} \bibinfo{person}{Ashish Sabharwal}.} \bibinfo{year}{2020}\natexlab{}.
\newblock \showarticletitle{Probing natural language inference models through semantic fragments}. In \bibinfo{booktitle}{\emph{Proceedings of the AAAI conference on artificial intelligence}}, Vol.~\bibinfo{volume}{34}. \bibinfo{pages}{8713--8721}.
\newblock


\bibitem[Sadrzadeh et~al\mbox{.}(2018)]%
        {sadrzadeh2018sentence}
\bibfield{author}{\bibinfo{person}{Mehrnoosh Sadrzadeh}, \bibinfo{person}{Dimitri Kartsaklis}, {and} \bibinfo{person}{Esma Balk{\i}r}.} \bibinfo{year}{2018}\natexlab{}.
\newblock \showarticletitle{Sentence entailment in compositional distributional semantics}.
\newblock \bibinfo{journal}{\emph{Annals of Mathematics and Artificial Intelligence}}  \bibinfo{volume}{82} (\bibinfo{year}{2018}), \bibinfo{pages}{189--218}.
\newblock


\bibitem[Samarinas et~al\mbox{.}(2020)]%
        {samarinas2020latent}
\bibfield{author}{\bibinfo{person}{Chris Samarinas}, \bibinfo{person}{Wynne Hsu}, {and} \bibinfo{person}{Mong~Li Lee}.} \bibinfo{year}{2020}\natexlab{}.
\newblock \showarticletitle{Latent retrieval for large-scale fact-checking and question answering with nli training}. In \bibinfo{booktitle}{\emph{2020 IEEE 32nd International Conference on Tools with Artificial Intelligence (ICTAI)}}. IEEE, \bibinfo{pages}{941--948}.
\newblock


\bibitem[Yusuf et~al\mbox{.}(2017)]%
        {yusuf2017basic}
\bibfield{author}{\bibinfo{person}{Aleshinloye~Abass Yusuf}, \bibinfo{person}{Nnanna~Agwu Nwojo}, {and} \bibinfo{person}{Moussa~Mahamat Boukar}.} \bibinfo{year}{2017}\natexlab{}.
\newblock \showarticletitle{Basic dependency parsing in natural language inference}. In \bibinfo{booktitle}{\emph{2017 13th International Conference on Electronics, Computer and Computation (ICECCO)}}. IEEE, \bibinfo{pages}{1--4}.
\newblock


\end{thebibliography}

\appendix

\section{Complementary Figures}

Below are complementary figures for our experiments and implementation. 
\begin{figure}[hbp]
    \centering
    \begin{minipage}{\columnwidth}
        \centering
            \includegraphics[width=\linewidth]{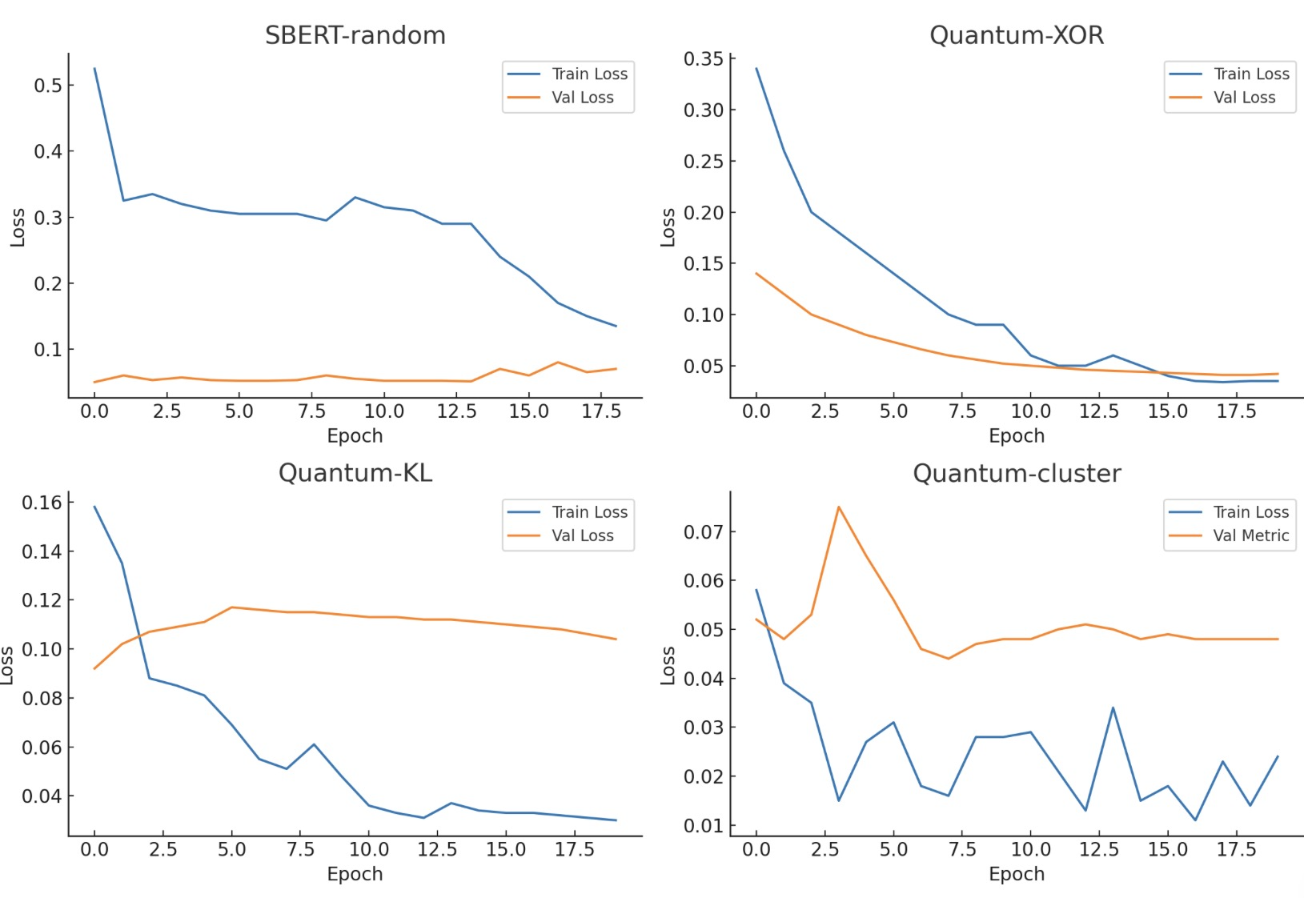}
    \caption{Training dynamics of quantum, hybrid, and transformer models on relatedness tasks. (a) SBERT-random model,
(b) Quantum-XOR model,
(c) Quantum-KL model,
(d) Quantum-cluster model.}\label{fig:related-train-comparison}
    \end{minipage}
\end{figure}

\begin{figure}[htbp]
    \centering
    \begin{minipage}{\columnwidth}
        \centering
            \includegraphics[width=\linewidth]{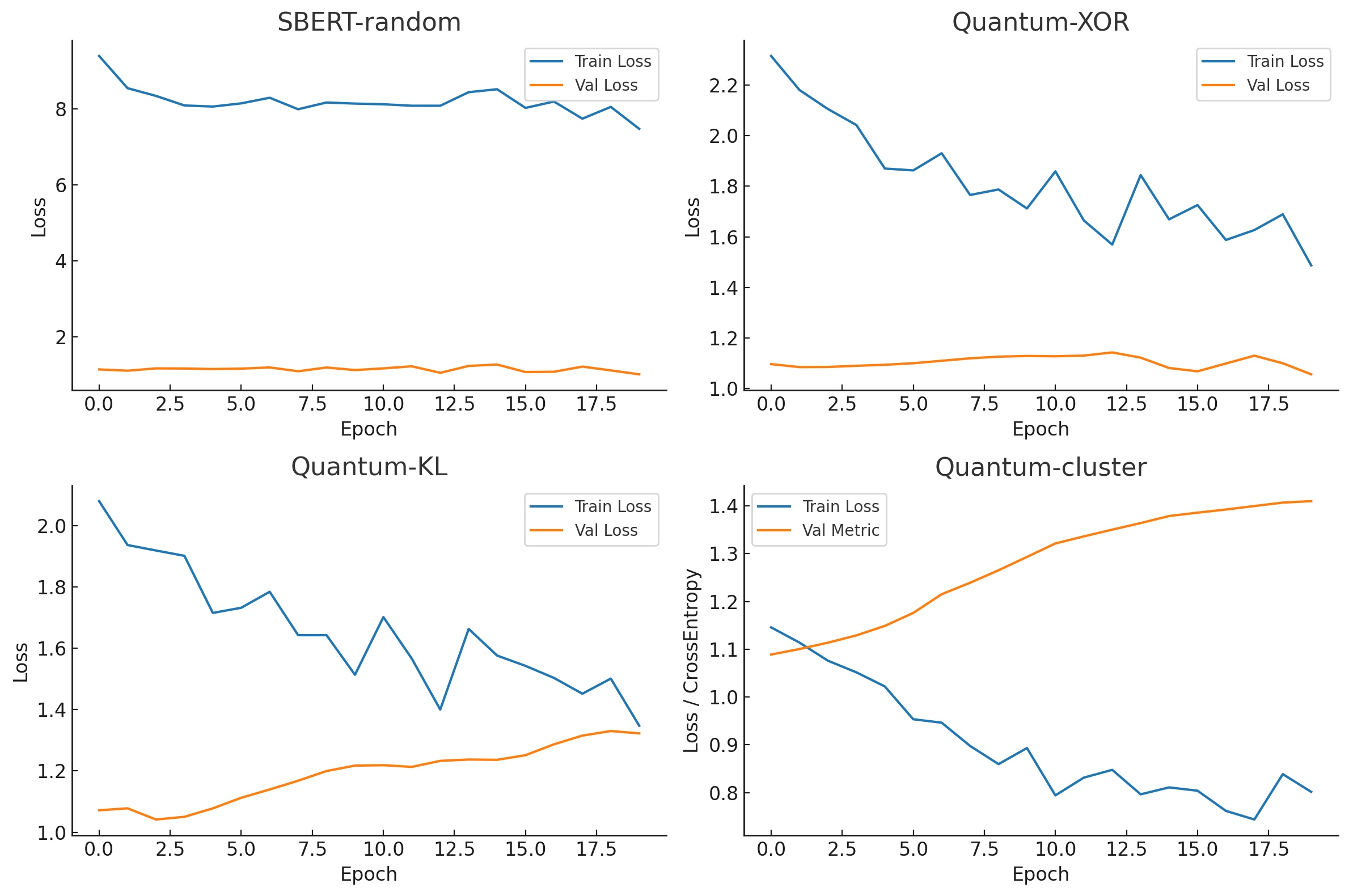}
    \caption{Training dynamics of quantum, hybrid, and transformer models on inference tasks. (a) SBERT-random model,
(b) Quantum-XOR model,
(c) Quantum-KL model,
(d) Quantum-cluster model.}\label{fig:inference-train-comparison}
    \end{minipage}
\end{figure}


\begin{figure}[htbp]
    \centering
    \begin{minipage}{\columnwidth}
        \centering
            \includegraphics[width=\linewidth]{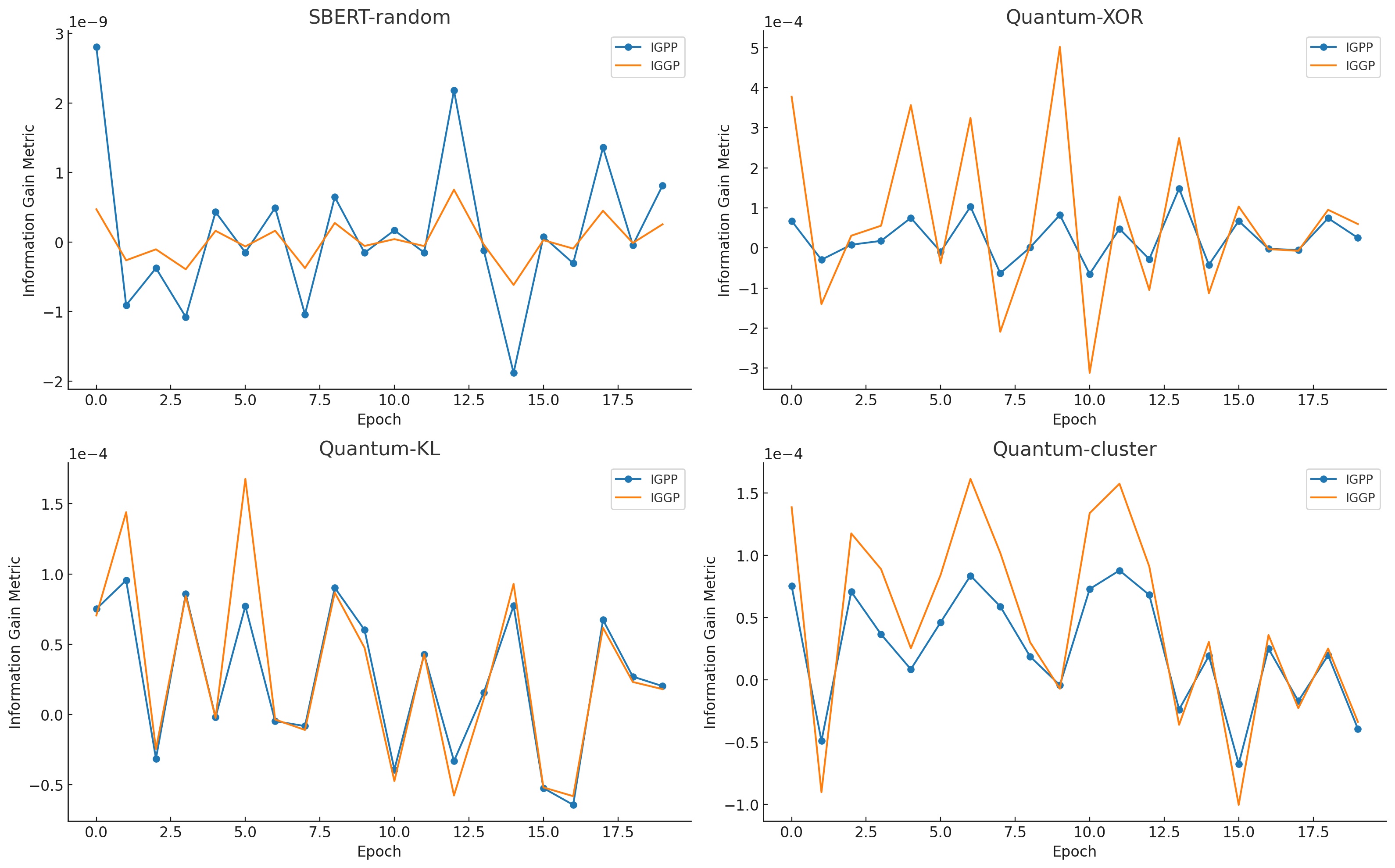}
    \caption{Epoch-wise Information Gain metrics during inference training across model types: (a) SBERT-random model,
(b) Quantum-XOR model,
(c) Quantum-KL model,
(d) Quantum-cluster model.}\label{fig:igpp-dynamics}
    \end{minipage}
\end{figure}

\end{document}